\documentclass[sigconf]{acmart}

\usepackage[table]{xcolor}
\usepackage{multirow}
\usepackage{booktabs} 
\usepackage{adjustbox}
\usepackage{xcolor}
\usepackage{threeparttable}
\usepackage{newunicodechar}
\usepackage{balance}
\newunicodechar{♫}{\textmusicalnote}
\theoremstyle{remark}
   
\newcommand{\colorCorrect}[1]{\textcolor[rgb]{0.3,0.6,0.3}{\textbf{#1}}}
\newcommand{\colorIncorrect}[1]{\textcolor[rgb]{0.7,0.3,0.3}{\textbf{#1}}}
\newcommand{\colorEvidence}[1]{\textcolor[rgb]{0.13,0.67,0.8}{\textbf{#1}}}
\newcommand{\token}[1]{\texttt{\small\textless#1\textgreater}}

\AtBeginDocument{%
  }

\copyrightyear{2026}
\acmYear{2026}
\setcopyright{cc}
\setcctype{by}
\acmConference[WSDM '26]{Proceedings of the Nineteenth ACM International Conference on Web Search and Data Mining}{February 22--26, 2026}{Boise, ID, USA}
\acmBooktitle{Proceedings of the Nineteenth ACM International Conference on Web Search and Data Mining (WSDM '26), February 22--26, 2026, Boise, ID, USA}
\acmPrice{}
\acmDOI{10.1145/3773966.3777986}
\acmISBN{979-8-4007-2292-9/2026/02}

\begin{document}

\title{CoDA: A Context-Decoupled Hierarchical Agent with Reinforcement Learning}

\author{Xuanzhang Liu}
\authornote{Equal contribution.}
\email{xuanzhangliu@gatech.edu}
\affiliation{%
  \institution{Georgia Institute of Technology}
  \city{Atlanta}
  \country{USA}
}

\author{Jianglun Feng}
\authornotemark[1]
\affiliation{%
  \institution{Alibaba Group}
  \city{Hangzhou}
  \country{China}
}

\author{Zhuoran Zhuang}
\authornotemark[1] 
\affiliation{%
  \institution{Alibaba Group}
  \city{Hangzhou}
  \country{China}
}

\author{Junzhe Zhao}
\affiliation{%
  \institution{Alibaba Group}
  \city{Hangzhou}
  \country{China}
}

\author{Maofei Que}
\authornote{Corresponding author.}
\affiliation{%
  \institution{Alibaba Group}
  \city{Hangzhou}
  \country{China}
}

\author{Jieting Li}
\affiliation{%
  \institution{Alibaba Group}
  \city{Hangzhou}
  \country{China}
}

\author{Dianlei Wang}
\affiliation{%
  \institution{Alibaba Group}
  \city{Hangzhou}
  \country{China}
}

\author{Hao Tong}
\affiliation{%
  \institution{Alibaba Group}
  \city{Hangzhou}
  \country{China}
}

\author{Ye Chen}
\affiliation{%
  \institution{Alibaba Group}
  \city{Hangzhou}
  \country{China}
}

\author{Pan Li}
\authornotemark[2]
\email{pan.li@scheller.gatech.edu}
\affiliation{%
  \institution{Georgia Institute of Technology}
  \city{Atlanta}
  \country{USA}
}

\renewcommand{\shortauthors}{Liu et al.}

\begin{abstract}
Large Language Model (LLM) agents trained with reinforcement learning (RL) show great promise for solving complex, multi-step tasks. However, their performance is often crippled by ``Context Explosion'', where the accumulation of long text outputs overwhelms the model's context window and leads to reasoning failures. To address this, we introduce CoDA, a Context-Decoupled hierarchical Agent, a simple but effective reinforcement learning framework that decouples high-level planning from low-level execution. It employs a single, shared LLM backbone that learns to operate in two distinct, contextually isolated roles: a high-level Planner that decomposes tasks within a concise strategic context, and a low-level Executor that handles tool interactions in an ephemeral, isolated workspace. We train this unified agent end-to-end using PECO (Planner-Executor Co-Optimization), a reinforcement learning methodology that applies a trajectory-level reward to jointly optimize both roles, fostering seamless collaboration through context-dependent policy updates. Extensive experiments demonstrate that CoDA achieves significant performance improvements over state-of-the-art baselines on complex multi-hop question-answering benchmarks, and it exhibits strong robustness in long-context scenarios, maintaining stable performance while all other baselines suffer severe degradation, thus further validating the effectiveness of our hierarchical design in mitigating context overload. Our code is available at \href{https://github.com/liuxuanzhang718/CoDA}{https://github.com/liuxuanzhang718/CoDA}.
\end{abstract}



\maketitle
\pagestyle{empty}    

\section{Introduction}

LLM agents have emerged as a new frontier in autonomous problem-solving, demonstrating their capability to tackle complex, multi-step tasks by interacting with external tools. To unlock their full potential, recent studies have focused on training these agents via reinforcement learning (RL) to autonomously learn reasoning and tool-use strategies \cite{Search-R1, R1-Searcher}. This approach, where agents learn from outcomes in interactive environments, holds the promise of creating truly adaptive and intelligent systems.

Despite their promising results, we identify a core limitation that lies in the current retrieval-augmented reasoning paradigm, which we term \textbf{Accumulated Context Explosion}. This refers to the phenomenon that \textit{as tasks grow in complexity, requiring more tool interactions, LLMs need to invoke external tools multiple times to query answers, which results in lengthy tool retrieval outputs}. This places significant demands on the agent’s context management, since previous approaches have primarily concatenated retrieval results directly into the context; however, the limited context window of current large models restricts their reasoning capabilities when dealing with lengthy inputs. For example, we have observed in our experiments that when each retrieval output reaches around 4000 tokens (see~\ref{exp:long_context}), the LLM model’s performance deteriorates significantly. Specifically, LLMs fail to access relevant information in the middle of long contexts\cite{lostinmiddle}. Even switching to models with larger context windows fails to alleviate this issue. 

\begin{figure}[t]
  \centering
  \includegraphics[width=\columnwidth]{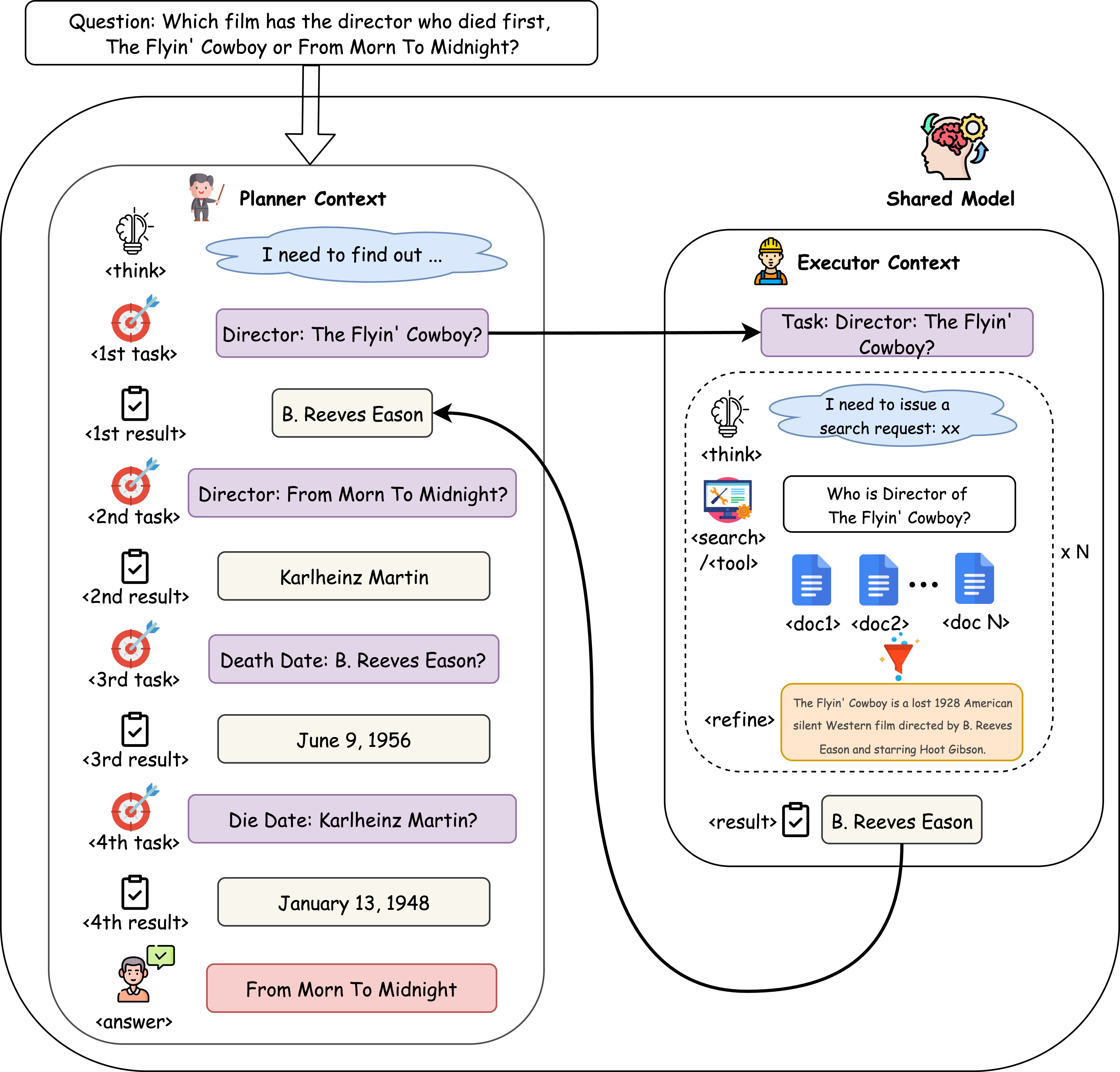}
  \caption{The workflow of CoDA. A single shared model acts as a Planner (left) to decompose the task and an Executor (right) to handle tool use in an isolated context.}
  \Description{Three line plots showing accuracy improvement over steps.}
  \label{fig:main_graph}
  \vspace{-0.4cm}
\end{figure}

Our core insight is that overcoming this cognitive overload requires a framework that not only structurally separates high-level planning from low-level execution, but is also holistically learnable. Inspired by the classic "divide and conquer" principle, we propose \textbf{CoDA} (Context-Decoupled Hierarchical Agent with Reinforcement Learning), a novel framework that we illustrate in Figure \ref{fig:main_graph}, which operationalizes this concept. It employs a single, shared LLM that learns to operate in two distinct, contextually-isolated roles:

\begin{enumerate}
    \item \textbf{Planner}: The Planner operates at a high level of abstraction, learning to decompose a complex request into a sequence of strategic, executable sub-tasks. It accepts the result from the Executor and plans the next step sequentially.
    \item \textbf{Executor}: For each sub-task, the same underlying model acts as the Executor. It functions within an ephemeral and isolated context to perform detailed reasoning, tool calls, and information synthesis related to the assigned sub-task.
\end{enumerate}

Critically, a \textbf{single, shared model} embodies both the Planner and Executor roles. This unified architecture enables \textbf{end-to-end training via reinforcement learning (RL)}, allowing for the coordinated optimization of both planning and execution and fostering seamless collaboration between the two roles.

We make the following contributions in this paper:
\begin{itemize}
    \item We propose CoDA, a novel hierarchical agent framework that mitigates context overload by decoupling high-level strategic planning from low-level task execution. This is achieved by maintaining role-specific contexts for a Planner and an Executor, embodied by a single, shared LLM.
    \item We introduce a Planner-Executor Co-Optimization (PECO) training methodology, an end-to-end reinforcement learning approach that jointly optimizes both roles. By applying a unified trajectory-level reward to context-dependent policy updates, PECO fosters seamless collaboration and mutual adaptation between the Planner and Executor.
    \item We achieve state-of-the-art performance on complex multi-hop QA benchmarks, as it significantly outperforms existing methods by as much as 6.0\% in terms of accuracy, demonstrating its superior reasoning capabilities.
    \item We demonstrate the superior long-context robustness of our framework. Through targeted experiments, we show that unlike baseline models whose performance degrades sharply with increasing context length, CoDA maintains stable and high performance, directly validating its effectiveness in handling information-intensive tasks.
\end{itemize}
\section{Related Work}

\subsection{Planner–Executor Agent Frameworks}
To overcome the limitations of monolithic agents in long-horizon tasks, the field has increasingly adopted Planner-Executor frameworks. This paradigm separates high-level strategic planning from low-level action execution. Recent work explore this concept through diverse strategies for decomposition and learning.

Some frameworks, like CoAct~\cite{CoAct} and PLAN-AND-ACT~\cite{erdogan}, decompose a complex goal into a sequence of natural language sub-tasks. Others, such as QCompiler~\cite{qcompiler}, employ a more formal, neuro-symbolic approach, "compiling" a query into a structured execution plan (an Abstract Syntax Tree). Architecturally, designs range from modular, multi-agent systems like AgentOrchestra~\cite{AgentOrchestra}, which uses specialized agents for different sub-tasks, to unified models. Prior approaches often design the planner and executor separately, which leads to potential mismatches between the two components and cannot be optimized as a unified system. As a result, executors may fail to follow the planner’s plans effectively, undermining overall task performance.

\subsection{Retrieval-Augmented Reasoning Agents}
Retrieval-Augmented Generation (RAG) has become a foundational technique for grounding Large Language Models (LLMs) in external, verifiable knowledge, thereby enhancing their factual accuracy and reducing hallucinations~\cite{borgeaud, RAG}. The evolution of this paradigm has evolved from simple, single-step retrieval mechanisms~\cite{gao2024retrievalaugmentedgenerationlargelanguage, FlashRAG} to more sophisticated, iterative pipelines designed to handle complex queries. These advanced RAG systems incorporate techniques such as query decomposition~\cite{qcompiler}, evidence refinement~\cite{RECOMP}, and multi-round search~\cite{shao-etal-2023-enhancing, trivedi}.

Despite these advances, traditional RAG methods often follow predefined workflows. This inherent rigidity can limit their ability to reason adaptively and make decisions dynamically when faced with novel or highly complex problems. To address this, a new class of agentic systems has emerged, integrating retrieval directly into the reasoning process as a learnable action \cite{R1-Searcher, Search-R1, ReSearch}. In these frameworks, the agent can autonomously decide when and what to search based on its intermediate reasoning steps.

\subsection{Context Engineering}

The effective performance of Large Language Model (LLM) agents is critically dependent on the quality and structure of the information they process. This has led to the formalization of Context Engineering \cite{surveycontextengineering}, which encompasses the systematic optimization of the LLM's informational payload, moving beyond simple prompt design. Our work is situated within this landscape, building upon foundational techniques and system-level architectures.

Research has identified the "lost-in-the-middle" problem, where models struggle to utilize information buried in long contexts~\cite{lostinmiddle}. In response, MemGPT~\cite{memgpt} pioneered OS-inspired memory management, creating a hierarchy between a limited active context (main memory) and external storage, which aligns with our principle of context separation. These components are increasingly integrated into multi-agent Systems (MAS), a paradigm that directly informs our hierarchical, role-based framework. Systems like AutoGen~\cite{autogen} and MetaGPT~\cite{metagpt} demonstrate how multiple agents, each with specialized roles and contexts, can collaborate to solve complex tasks. This approach embodies the principles of task specialization and context isolation, where each agent focuses on its own sub-problem. 

\subsection{RL for Agents}

The paradigm for enhancing the capabilities of LLMs as agents is increasingly shifting from supervised fine-tuning (SFT) on curated datasets~\cite{toolformer, ToolLLM}, towards the more scalable and generalizable framework of RL. End-to-end RL offers a promising alternative to unlock the inherent reasoning capabilities of LLMs, with recent breakthroughs demonstrating its effectiveness\cite{deepseekr1}. The development of RL methods has progressed from foundational techniques like Reinforcement Learning from Human Feedback (RLHF) \cite{RLHF} and Proximal Policy Optimization (PPO)~\cite{PPO} to more advanced approaches such as Direct Preference Optimization (DPO) \cite{DPO}, and Group Relative Policy Optimization (GRPO)~\cite{GRPO}. Among these methods, GRPO~\cite{GRPO} is particularly relevant as it is specifically designed for LLMs. By replacing the traditional critic with a group-based evaluation strategy, GRPO has demonstrated strong performance in enhancing reasoning across tasks. In the domain of external knowledge retrieval, systems such as Search-R1~\cite{Search-R1}, ReSearch~\cite{ReSearch}, and R1-Searcher~\cite{R1-Searcher, R1-Searcher++} have successfully leveraged RL to enable models to autonomously develop reasoning strategies during the retrieval process, moving beyond manually specified cues. Building upon this proven efficacy and adaptability, our work extends GRPO to enhance retrieval-augmented reasoning, significantly improving the ability of LLMs to interact with external tools in question answering (QA) tasks. Furthermore, we further extend GRPO within a hierarchical agent framework, addressing the unique challenges of multi-step reasoning scenarios.

\subsection{Hierarchical Reinforcement Learning (HRL)}
\label{sec:related_hrl}

Hierarchical Reinforcement Learning (HRL) has been extensively studied to decompose long-horizon tasks into simpler subproblems \cite{hrl-survey}. Foundational frameworks include \textit{Options} \cite{sutton}, which models temporal abstraction through semi-Markov decision processes, and the \textit{MAXQ} value function decomposition \cite{MAXQ}. Of particular relevance to our work is the \textit{Feudal Reinforcement Learning} paradigm introduced by Dayan and Hinton \cite{Feudal}, which employs a "Manager-Worker" hierarchy where high-level policies set goals for low-level agents to execute. This structure has been successfully adapted to Deep RL settings, such as h-DQN \cite{temporal} and FeUdal Networks (FuN) \cite{feudalhrl}, to address challenges in exploration and sparse reward environments \cite{nachum2018dataefficienthierarchicalreinforcementlearning}.

While CoDA draws structural inspiration from these Feudal hierarchies, our motivation and application differ fundamentally. Traditional HRL primarily focuses on \textit{temporal abstraction} to solve the "curse of dimensionality" in state spaces and enable efficient exploration over long time horizons. In contrast, CoDA applies hierarchical principles to address the unique \textit{contextual constraints} of Large Language Models. Instead of solving for sparse rewards, our "Planner-Executor" separation is designed to mitigate \textit{Context Explosion} and \textit{Context Entanglement}, ensuring that the high-level planner remains isolated from the noisy, verbose outputs generated during low-level execution. Thus, CoDA repurposes the HRL architecture for informational decoupling rather than purely for temporal credit assignment.
\section{Method}

In this section, we first formalize the problem and core challenges, and then introduce our proposed \textbf{CoDA} framework to address these challenges. Finally, we describe the RL strategy employed to train this framework, detailing our context-dependent policy updates and composite reward function.

\subsection{Problem Formulation}
\subsubsection{Context Entanglement in Monolithic Agents}
We address the task of complex, multi-step question answering where an agent, parameterized by a Large Language Model $\pi_\theta$, interacts with a set of external tools $T$ to answer a query $Q$. A common approach models this as a sequential decision-making process, where the agent generates a single, continuous trajectory $\tau = (y_1, \ldots, y_L)$. 

In this standard paradigm, the context $C_t$ provided to the policy $\pi_\theta(y_t \mid C_t)$ at each step $t$ is a monolithic concatenation of all prior history: $C_t = [Q; y_1; \ldots; y_{t-1}]$. As the reasoning chain lengthens with multiple tool calls, this context rapidly accumulates raw, verbose documents $(D_1, \ldots, D_k)$, leading to two widely recognized issues: computational inefficiency due to processing long sequences, and performance degradation as critical information gets ``lost in the middle''\cite{lostinmiddle} of an overwhelmingly long input.

We argue that these issues are symptoms of a deeper, structural flaw inherent in the monolithic design. Complex reasoning naturally comprises two distinct cognitive levels: high-level strategic planning (decomposing the problem) and low-level tactical execution (fulfilling a specific sub-task). This paradigm forces these two roles to operate within the same context, creating a state of \emph{Context Entanglement}, which manifests as two primary challenges:

\begin{itemize}
    \item \textbf{Strategic Context Pollution:} The Planner's decision space, which should ideally focus on high-level goals and results, is contaminated by the noisy, low-level details of raw documents. This obscures the strategic signals required for effective long-range planning.
    \item \textbf{Execution Context Redundancy:} When the agent acts as an Executor for a specific task$_t$, it is burdened with a context $C_t$ containing the entire, often irrelevant, history of previous tasks ($\text{task}_1$, $\dots$, $\text{task}_{t-1}$). This not only dramatically inflates the context length, consuming valuable tokens and increasing latency, but also introduces unrelated information that can interfere with the focused execution of the current task.
\end{itemize}

Formally, the agent's objective is to find the optimal parameters $\theta^*$ that maximize the expected cumulative reward $R(\tau)$ over trajectories sampled from the policy:
\begin{equation}
    \theta^* = \arg\max_\theta \mathbb{E}_{\tau \sim \pi_\theta} \left[R(\tau)\right]
\end{equation}
However, due to the dual challenges of strategic pollution and context redundancy stemming from cognitive entanglement, the policy's effectiveness $\pi_\theta(y_t \mid C_t)$ diminishes as task complexity and context length $|C_t|$ grow. 

A seemingly intuitive approach to mitigate this context explosion would be to programmatically remove or truncate verbose tool outputs (e.g., the documents $D_k$) from the history $C_t$ after they have been processed. However, naively truncating tool outputs at inference can introduce a distribution shift if the model was never trained under matching context dynamics. This introduces a severe \textbf{Training-Inference Skew}: the model is trained to generate each token conditioned on the full, unaltered history, but during inference, it would be forced to operate on an artificially shortened context it has never encountered during training. This distribution shift cripples the model's ability to reason coherently, as its learned policy is no longer valid in this modified state space. This demonstrates that a simple post-hoc context truncation is not a viable solution. The problem of context entanglement requires a more principled, architectural redesign. Our work seeks to resolve this by fundamentally redesigning the agent's operational framework to decouple these entangled roles.

\subsection{Hierarchical Framework}
To address the challenge, we propose the CoDA framework. Its core idea is to decompose the monolithic decision-making process into two synergistic yet contextually-isolated logical roles: a \textbf{Planner} and an \textbf{Executor}. Both roles are embodied by a \textbf{single} model $\pi_\theta$ but operate under distinct contexts.

\subsubsection{Role Definitions and Context Spaces}
\begin{description}
    \item[The Planner] acts as a high-level strategist, operating within a concise \textbf{Strategic Context} $C_P$. Its responsibility is to generate sub-tasks or the final answer. The strategic context is kept clean, containing only high-level information:
    \[
    C_P^{(t)} = \{Q, (\text{task}_1, \text{result}_1), \dots, (\text{task}_{t-1}, \text{result}_{t-1})\}
    \]
    At each step $t$, the Planner generates the next sub-task, $\text{task}_t = \pi_{\theta}(C_P^{(t)})$. The process terminates when the Planner determines that it has gathered enough information. At this point, it generates a final action, which is to synthesize the collected results into the final answer. This final synthesis can be viewed as the last "task" in the sequence.

    \item[The Executor] acts as a focused task handler, operating within a \textbf{Temporary Execution Context} $C_E$. It receives a single $\text{task}_t$ from the Planner and is responsible for its fulfillment through tool use and information processing. Its internal workflow is as follows:
    \begin{enumerate}
        \item \textbf{Initialize:} $C_E \leftarrow \{\text{task}_t\}$.
        \item \textbf{Execute:} Invoke tools (e.g., $\mathcal{T}_{\text{search}}$) to retrieve raw documents $D_t$.
        \item \textbf{Refine:} Summarize and distill $D_t$ to produce a concise information summary $s_t = \text{Refine}(D_t)$.
        \item \textbf{Conclude:} Generate a final, condensed $\text{result}_t$ based on the refined information within $C_E$. This $\text{result}_t$ is a concise summary of the findings for $\text{task}_t$, ready to be passed back to the Planner.
    \end{enumerate}
\end{description}
Crucially, the verbose, raw documents $D_t$ never enter the Planner's strategic context $C_P$. The Executor functions as a black box that abstracts away noisy details, enabling effective context decoupling.

\subsection{PECO (Planner-Executor Co-Optimization)}

We train the CoDA framework end-to-end using an outcome-supervised reinforcement learning approach based on the GRPO algorithm \cite{GRPO}. Our core strategy hinges on applying a single trajectory-level reward to update the shared policy model $\pi_{\theta}$, while ensuring the gradient for each token is calculated based on its specific role-dependent context (Planner or Executor).

\subsubsection{Hierarchical Trajectory Generation (Rollout)}
Training begins by generating interaction trajectories using the behavior policy $\pi_{\theta_{\text{old}}}$. The standard rollout is illustrated as follows:
\begin{enumerate}
    \item \textbf{Planner Action:} Starting with the user query, the Planner generates a sequence of thoughts and an action, which is either a \texttt{<task>} to delegate or a final \texttt{<answer>}.
    \item \textbf{Executor Sub-Loop:} If the Planner generates a \texttt{<task>}, we invoke a separate generation loop. This session operates in an ephemeral context, initialized only with the sub-task, cutting the session off from the Planner's long history. Within this ephemeral context, the model, now acting as the Executor, focuses solely on the sub-task. It can iteratively generate \texttt{<search>} actions, receive external tool outputs (e.g., \texttt{<documents>...</documents>}), and reason over them until it produces an \texttt{<answer>} for the sub-task.
    \item \textbf{Returning to Planner:} The Executor's final answer is packaged as a \texttt{<result>} and returned to the Planner, becoming part of the Planner's strategic context for its next decision-making step.
    \item \textbf{Trajectory Collection:} All generated sequences—the main Planner trajectory and all associated Executor trajectories—are collected. They are then concatenated into a single training batch, where each trajectory is explicitly labeled as either 'planner' or 'executor' to facilitate group-level credit assignment. This collection of trajectories, representing a complete multi-level reasoning process for a single query, forms a \textbf{trajectory group}.
\end{enumerate}

\subsubsection{Group-Level Credit Assignment}
For each query, we generate $k$ > 1 independent trajectories, termed as 'rollouts', using the same policy. All trajectories (Planner and Executor) originating from the same query form a trajectory group $\mathcal{G}$. We compute a scalar reward $R(\mathcal{G})$ for each $\mathcal{G}$ based on the function detailed in Section~\ref{subsubsec:reward_design}.

Following GRPO\cite{GRPO} principles, we assign credit by normalizing the reward of each group against the statistics of its peers that tackle the same initial query. If the $k$ rollouts for a single query yield a set of rewards 
$\{R(\mathcal{G}_1), R(\mathcal{G}_2), \dots, R(\mathcal{G}_k)\}$, the advantage 
$\hat{A}(\mathcal{G}_i)$ for the $i$-th group is calculated as:

\begin{equation} 
\hat{A}(\mathcal{G}_i) = \frac{R(\mathcal{G}_i) - \text{mean}({ R(\mathcal{G}_j) }_{j=1}^k)}{\sigma({ R(\mathcal{G}_j) }_{j=1}^k)}
\end{equation}
where $\text{mean}$ and $\sigma$ are the mean and std computed exclusively over the $k$ outcomes for that specific query. This advantage is applied uniformly to every policy-generated token across all trajectories (both Planner and Executor) within that group $\mathcal{G}_i$. This outcome-based approach ensures that even intermediate Executor actions are credited based on their contribution to the Planner's final success.

\subsubsection{Context-Dependent Policy Update with Loss Masking}
A cornerstone of our training strategy is ensuring the model learns exclusively from its own \textbf{actions}, not from \textbf{observations} provided by the environment. While the advantage signal $\hat{A}(\mathcal{G}_i)$ is uniform, the policy update is performed token-by-token, strictly enforcing this principle through loss masking.

We use the Group Relative Policy Optimization (GRPO) objective to update the policy $\pi_{\theta}$ against the behavior policy $\pi_{\theta_{\text{old}}}$. For each token $y_t$ generated by $\pi_{\theta}$ within a group $\mathcal{G}_i$, the loss is:

\begin{equation}
\label{eq:clip_loss}
\small{L_{\text{CLIP}}(y_t, C_t, \theta) = \min \left( \rho_t(\theta) \hat{A}(\mathcal{G}_i), \operatorname{clip}(\rho_t(\theta), 1-\epsilon, 1+\epsilon) \hat{A}(\mathcal{G}_i) \right)}
\end{equation}
where $\rho_t(\theta) = \frac{\pi_\theta(y_t | C_t)}{\pi_{\theta_{\text{old}}}(y_t | C_t)}$ is the importance sampling ratio.

The overall learning objective for the policy $\pi_\theta$ is to maximize the GRPO surrogate objective function $J(\theta)$, regularized by a KL divergence term against a reference policy $\pi_{\theta_{\text{ref}}}$:

\begin{equation}
\label{eq:grpo_objective}
\scriptscriptstyle
J(\theta) = \mathbb{E}_{\mathcal{G} \sim \pi_{\theta_{\text{old}}}} \Biggl[ \sum_{\tau \in \mathcal{G}} \sum_{t=0}^{L_{\tau}-1} \Biggl(L_{\text{CLIP}}(y_t, C_t, \theta) \nonumber - \beta D_{KL}\left(\pi_{\theta}(y_t | C_t) \parallel \pi_{\theta_{\text{ref}}}(y_t | C_t)\right) \Biggr) \cdot m_t \Biggr]
\end{equation}

The mask $m_t$ is the crucial component that enforces the action-observation separation. During trajectory construction, our framework distinguishes between tokens generated by the agent and tokens returned by the environment. The loss is computed only on agent-generated tokens by applying a binary loss mask $m_t$:
\begin{itemize}
    \item Agent's Actions ($m_t = 1$): tokens are generated by the agent as part of its decision-making process. This includes the Planner's reasoning and \texttt{<task>} generation, as well as the Executor's reasoning and \texttt{<search>} generation.
    \item Environmental Observations ($m_t = 0$): tokens represent information from the environment. For the Planner, the output in the \texttt{<result> \dots </result>} block is an observation. For the Executor, the search result in the \texttt{<documents> \dots </documents>} block is also an observation. 
\end{itemize}

\subsubsection{Composite Reward Design}
\label{subsubsec:reward_design}

To guide the model towards mastering the complex behaviors required by our framework, we designed a composite reward function $R(\mathcal{G})$ that simultaneously encourages three distinct objectives: answer correctness, structural compliance, and information refinement quality, where total reward for a complete trajectory group $\mathcal{G}$ is a weighted sum of them.

\paragraph{1. Answer Correctness ($R_{\text{ans}}$).}\label{reward_design_1}
The primary objective is to produce a factually correct answer.  This reward is based on the F1 score between the final answer generated by the Planner $A_{\text{pred}}$, and the set of ground-truth answers $G$. We linearly transform this score to a range of $[-3, 3]$ to provide a strong, scaled reward signal.
\begin{align}
S_{\text{ans}} &= \max_{g \in G} \text{F1}(A_{\text{pred}}, g) \\
R_{\text{ans}}(\mathcal{G}) &= 6 \cdot S_{\text{ans}} - 3
\end{align}

\paragraph{2. Format Compliance ($R_{\text{format}}$).}\label{reward_design_2}
To ensure the agent adheres to the Plan-then-Execute protocol, we reward the generation of correctly formatted XML-style tags. This component is composed of two parts, one for the Planner and one for the Executor.
\begin{equation}
R_{\text{format}}(\mathcal{G}) = I_P(\mathcal{G}) + I_E(\mathcal{G})
\end{equation}
Here, $I_P(\mathcal{G})$ is an indicator function that equals 1 if the Planner's final output in the group is structurally correct (i.e. proper use of \texttt{<task>} or \texttt{<answer>}). Similarly, $I_E(\mathcal{G})$ is an indicator function that equals 1 if all of the Executor's outputs within the group are correctly formatted (i.e. proper use of \texttt{<search>} or \texttt{<result>})

\paragraph{3. Refinement Quality ($R_{\text{refine}}$).}\label{reward_design_3}
A crucial capability of the Executor is to distill critical information from noisy search results. To incentivize not only the act of summarization but also its factual accuracy, the refinement reward is granted only if two conditions are met: a non-empty summary is produced, and this summary contains the ground-truth answer. 

Let $s_{\text{refine}}(\tau_e)$ be the set of all string contents extracted from \texttt{<refine>} tags across all Executor trajectories within a group $\mathcal{G}$. We first concatenate these contents into a single, unified string, $s_{\text{combined}}$. 
\begin{equation}
s_{\text{combined}} = \bigoplus_{i=1}^{m} s_i = s_1 \oplus s_2 \oplus \dots \oplus s_m
\end{equation}
where $\oplus$ denotes the string concatenation operator. The refine reward is formally defined as:
\begin{equation}
R_{\text{refine}}(\mathcal{G}) = \delta \cdot \mathbb{I}(s_{\text{combined}} \neq \emptyset) \cdot \mathbb{I}(G \subseteq s_{\text{combined}})
\end{equation}
where $\mathbb{I}(\cdot)$ is the indicator function, the condition $G \subseteq s_{\text{combined}}$ checks for the textual containment of at least one ground-truth answer from the set $G$within the combined refined string, and $\delta$ is its corresponding weight. This approach rewards the collective refined effort of the Executor, even if critical information is distributed across multiple refinement steps.

\paragraph{\textbf{Total Reward}}
The final reward for a trajectory is the sum of these components:
\begin{equation}
R(\mathcal{G}) = R_{\text{ans}}(\mathcal{G}) + R_{\text{format}}(\mathcal{G}) + R_{\text{refine}}(\mathcal{G})
\end{equation}
This multi-faceted reward function effectively provides dense signals that guide the agent to concurrently optimize for accuracy, procedural correctness, and semantic understanding.

\vspace{-0.5em} 
\subsection{Efficiency Analysis}
\label{sec:appendix_complexity}

To demonstrate CoDA's computational advantages, we analyze its time and space complexity ($T(C)=\mathcal{O}(C^2)$, $S(C)=\mathcal{O}(C)$ for context length $C$) against a monolithic agent.
Let $H$ be the number of high-level planning hops, $h$ the executor's internal hops (usually small, e.g., $h=1$), $L_{\text{doc}}$ the raw document length, and $L_{\text{res}}$ the concise result length. Crucially, $L_{\text{res}} \ll L_{\text{doc}}$. Other terms like query length $L_Q$ are treated as lower-order constants.

\textbf{Monolithic Agent.} A monolithic agent accumulates all history in a single context. At step $H$, the peak context is $C_{\text{mono}}^{\text{max}} \approx H \cdot L_{\text{doc}}$. This results in multiplicative growth: space complexity $S_{\text{mono}} = \mathcal{O}(H \cdot L_{\text{doc}})$ and quadratic time complexity $T_{\text{mono}} = \mathcal{O}((H \cdot L_{\text{doc}})^2)$, leading to intractable "context explosion."

\textbf{CoDA (Hierarchical).} CoDA decouples context into two isolated roles.
The \textit{Executor} operates in an ephemeral workspace reset for each sub-task, bounded by a single document: $C_{\text{exec}}^{\text{max}} \approx \mathcal{O}(h \cdot L_{\text{doc}})$, independent of $H$.
The \textit{Planner} maintains a strategic context of concise summaries: $C_{\text{plan}}^{\text{max}} \approx H \cdot L_{\text{res}}$.
The system's peak complexity depends on the maximum of these two independent contexts, rather than their product. Thus, CoDA's space complexity is $S_{\text{CoDA}} = \mathcal{O}(\max(H \cdot L_{\text{res}}, h \cdot L_{\text{doc}}))$ and time complexity is $T_{\text{CoDA}} = \mathcal{O}(\max((H \cdot L_{\text{res}})^2, (h \cdot L_{\text{doc}})^2)$.
Since $L_{\text{res}} \ll L_{\text{doc}}$, CoDA effectively prevents large documents from accumulating in the planner's context, ensuring linear scalability with respect to task hops $H$.

\section{Experiments}
In this section, we conduct a series of experiments to evaluate the effectiveness of CoDA, and to answer the following research questions:
\begin{enumerate}
    \item How does CoDA compare against state-of-the-art baselines on single-hop and multi-hop QA benchmarks?
    \item What is the impact of hierarchical architecture and composite reward function in CoDA?
    \item How robust CoDA in long-context scenarios?

\end{enumerate}
\subsection{Experimental Setup}

\subsubsection{Datasets}
We use the following seven widely used QA datasets, covering both single-hop and multi-hop reasoning. 
\begin{itemize}
    \item \textbf{Single-Hop QA:} Natural Questions (NQ)~\cite{nq}, TriviaQA~\cite{triviaqa}, and PopQA~\cite{popqa}.
    \item \textbf{Multi-Hop QA:} HotpotQA~\cite{hotpotqa}, 2WikiMultiHopQA~\cite{2WikiMultiHopQA}, Musique~\cite{musique}, and Bamboogle~\cite{bamboogle}.
\end{itemize}
To simulate a real-world open-domain setting, we remove all provided context documents.

\subsubsection{Evaluation Metrics}
We evaluate model performance using three standard metrics:
\begin{itemize}
\item \textbf{Exact Match (EM):} \% of predictions that exactly match one of the ground-truth answers after normalization (e.g., lowercasing, removing punctuation and articles).
\item \textbf{F1-Score:} The harmonic mean of precision and recall computed at the token level, which measures the overlap between the prediction and ground-truth answers.
\item \textbf{CEM (Coverage Exact Match):} A more tolerant version of EM that scores 1 if a normalized ground-truth answer is a substring of the normalized prediction.
\end{itemize}

\subsubsection{Baselines}
We compare our CoDA against a range of strong baselines, including both closed-source and open-source models.
\begin{itemize}
\item \textbf{Closed-Source Models:} We include results from GPT-4o and GPT-4o-mini~\cite{gpt4o}, representing state-of-the-art proprietary language models.
\item \textbf{Open-Source Methods:} We evaluate several reasoning and search-augmentation methods built upon open-source LLMs. Most baseline results are sourced directly from Search-R1~\cite{Search-R1}, which shares our experimental setup. These methods include Naive Response, ReAct \cite{react}, Search-o1, Search-R1, and AutoRefine~\cite{Autorefine}. All open-source methods are evaluated with the Qwen2.5-3B-Base backbone for fairness.
\end{itemize}

\subsubsection{Implementation Details} In all experiments, models perform search over the December 2018 Wikipedia dump~\cite{wiki18}, which serves as the external knowledge corpus. We use E5-base-v2~\cite{e5basev2} as our retriever. For each search query, the retriever returns the top-3 most relevant documents. Our reinforcement learning (RL) framework is implemented using verl~\cite{verl}. We employ Group Relative Policy Optimization (GRPO)~\cite{GRPO} as the learning algorithm to train our agent, which is based on Qwen2.5-3B-Base.

\subsection{Main Results (RQ1)}
Table~\ref{tab:exp_main} presents the main results of our CoDA framework compared against a range of baseline methods on seven diverse QA benchmarks. Our proposed model, CoDA-Base, demonstrates state-of-the-art performance, achieving the highest average score of 0.407 across all datasets.

On single-hop QA tasks (NQ, TriviaQA, PopQA), which require less reasoning depth, CoDA performs only marginally better than the strong AutoRefine baseline, improving by approximately 2\% on average. This is expected, as these tasks do not extensively challenge the model’s context management capabilities, and thus the primary advantage of our framework is less pronounced.

The true strength of the CoDA framework becomes evident in complex, multi-hop QA scenarios (HotpotQA, 2WikiMultiHopQA, Musique, Bamboogle). On these challenging benchmarks, CoDA consistently and significantly outperforms all baselines. The most notable gains are observed on 2WikiMultiHopQA and Musique—two notoriously difficult datasets requiring intricate reasoning over multiple documents. Here, CoDA surpasses the strong AutoRefine baseline by a remarkable 24\% on 2WikiMultiHopQA and 10\% on Musique.

This huge improvement on multi-hop tasks directly stems from CoDA's design: hierarchical task decomposition and contextual separation. In monolithic agent architectures, the reasoning trace becomes progressively entangled with raw, unfiltered search results from multiple retrieval steps, where the model's high-level plan is intermingled with low-level, noisy data. CoDA addresses this by separating the roles of the planner and executor, where the planner’s trajectory is shielded from raw search results.

\begin{table}[t]
    \centering
    \small
    \caption{
    \textbf{(RQ1)}
    EM comparison of CoDA versus baseline methods with Qwen2.5-3B \cite{qwen25} across various QA benchmarks.
    \textbf{Bold} denotes best results; \underline{underline} denotes the second-best; $***$ indicates significance at the 0.01 level.
    }
    \resizebox{\linewidth}{!}{
    \begin{tabular}{lllllllll}
        \toprule
        & \multicolumn{3}{c}{Single-Hop QA} & \multicolumn{4}{c}{Multi-Hop QA}       & \multicolumn{1}{l}{} \\
        \cmidrule(lr){2-4} \cmidrule(lr){5-8}
        Methods & NQ & TriviaQA & PopQA$^*$ & HotpotQA  & 2Wiki & Musique & Bamboogle & Avg. \\

        \midrule
        \multicolumn{9}{l}{{w/o Retrieval}} \\
        \quad Direct Generation     & 0.106   & 0.288      & 0.108   & 0.149    & 0.244 & 0.020   & 0.024     & 0.134 \\
        \quad SFT                  & 0.249   & 0.292      & 0.104   & 0.186    & 0.248 & 0.044   & 0.112     & 0.176 \\
        \quad R1-Instruct \cite{deepseekr1}          & 0.210   & 0.449      & 0.171   & 0.208    & 0.275 & 0.060   & 0.192     & 0.224 \\
        \quad R1-Base \cite{deepseekr1}              & 0.226   & 0.455      & 0.173   & 0.201    & 0.268 & 0.055   & 0.224     & 0.229 \\
        \midrule
        \multicolumn{9}{l}{{w/ Single-Hop Retrieval}} \\
        \quad Naive RAG \cite{RAG}                  & 0.348   & 0.544      & 0.387   & 0.255    & 0.226 & 0.047   & 0.080     & 0.270 \\
        \midrule
        \multicolumn{9}{l}{{w/ Multi-Hop Retrieval}} \\
        \quad Search-o1 \cite{Search-o1}                 & 0.238   & 0.472      & 0.262   & 0.221    & 0.218 & 0.054   & 0.320     & 0.255 \\
        \quad IRCoT \cite{ircot} & 0.111   & 0.312      & 0.200   & 0.164    & 0.171 & 0.067   & 0.240     & 0.181 \\
        \quad ReSearch-Base \cite{ReSearch}             & \underline{0.427}          & \underline{0.597}          & 0.430 & 0.305 & 0.272 & 0.074 & 0.128 & 0.319 \\
        \quad Search-R1-Base \cite{Search-R1}           & 0.421   & 0.583      & 0.413   & 0.297    & 0.274 & 0.066   & 0.128     & 0.312 \\
        \quad AutoRefine-Base  \cite{Autorefine}                 
        & 0.424 & 0.587 & \underline{0.449} & \underline{0.382} & \underline{0.328} & \underline{0.169} & \underline{0.320} & \underline{0.380} \\
        \midrule
        \quad \textbf{CoDA-Base}                  
        & \textbf{0.441} & \textbf{0.608***} &\textbf{0.452} & \textbf{0.407***} & \textbf{0.408***} & \textbf{0.187} & \textbf{0.368} & \textbf{0.407***} \\
        \quad \textbf{Improve \%} & \textbf{+4.01\%} & \textbf{+3.58\%} & \textbf{+0.67\%} & \textbf{+6.54\%} & \textbf{+24.39\%} & \textbf{+10.65\%} & \textbf{+15.00\%} & \textbf{+7.11\%} \\  
        \bottomrule
        \end{tabular}
    }
    \parbox{\linewidth}{
    \raggedright
    \scriptsize
    $^*$  PopQA contains duplicate queries with different gold answers, so we unify answers per query and re-evaluate.
    \par}
    \label{tab:exp_main}
\end{table}

\subsection{Ablation Study (RQ2)}

\subsubsection{CoDA w/o Hierarchical Structure}

To empirically validate that our hierarchical design is the primary driver of performance, we conducted an ablation study comparing agents with and without this structure. We compared monolithic version of CoDA, (i.e. AutoRefine baseline). Additionally, to demonstrate the generalizability of the architectural benefit, we applied our hierarchical structure to the Search-R1 baseline. Table~\ref{tab:exp_ablation_hierarchical} provides compelling evidence for the superiority of our design. Applying the structure to the standard Search-R1 model boosts its average EM score from 0.312 to 0.375, with the most dramatic gains on complex multi-hop datasets like Musique. Similarly, our full CoDA model significantly outperforms its monolithic counterpart (0.407 vs. 0.380).

\begin{table}[t]
    \centering
    \caption{
    Ablation study comparing Search-R1 and CoDA with/without hierarchical structure. All values are EM scores.
    }
    \resizebox{\linewidth}{!}{
    \begin{tabular}{lcccccccc}
        \toprule
        & \multicolumn{3}{c}{Single-Hop QA} & \multicolumn{4}{c}{Multi-Hop QA}       & \multicolumn{1}{l}{} \\
        \cmidrule(lr){2-4} \cmidrule(lr){5-8}
        Model Variants            & NQ      & TriviaQA   & PopQA   & HotpotQA & 2wiki & Musique & Bamboogle & Avg.                 \\
        \midrule
        \rowcolor[gray]{0.9}
        Search-R1 &  &  &  &  &  &  &  & \\ 
        \quad{w} Hier. & \textbf{0.441} & \textbf{0.601} & \textbf{0.423} & \textbf{0.388} & \textbf{0.284} & \textbf{0.151} & \textbf{0.339} & \textbf{0.375} \\ 
        \quad{w/o} Hier. & 0.421 & 0.583 & 0.413 & 0.297 & 0.274 & 0.066 & 0.128 & 0.312 \\
        \midrule
        \rowcolor[gray]{0.9}
        CoDA &  &  &  &  &  &  &  & \\ 
        \quad{w} Hier. & \textbf{0.441} & \textbf{0.608} &\textbf{0.452} & \textbf{0.407} & \textbf{0.408} & \textbf{0.187} & \textbf{0.368} & \textbf{0.407}\\ 
        \quad{w/o} Hier. & 0.424 & 0.587 & 0.449 & 0.382 & 0.328 & 0.169 & 0.320 & 0.380\\
        \bottomrule
        \end{tabular}
    }
    \label{tab:exp_ablation_hierarchical}
\end{table}

\subsubsection{Planner \& Executor Co-optimization(PECO)}

To isolate and validate the contribution of our reinforcement learning methodology, we conducted an ablation study comparing the full CoDA framework against a version that relies solely on prompting for its hierarchical behavior. We established a strong workflow baseline where the agent utilizes the same Planner-Executor architecture, but its actions are guided exclusively by the same carefully designed in-context prompts, with no RL-based parameter updates. 

\begin{figure}[t]
  \centering
  \includegraphics[width=\columnwidth]{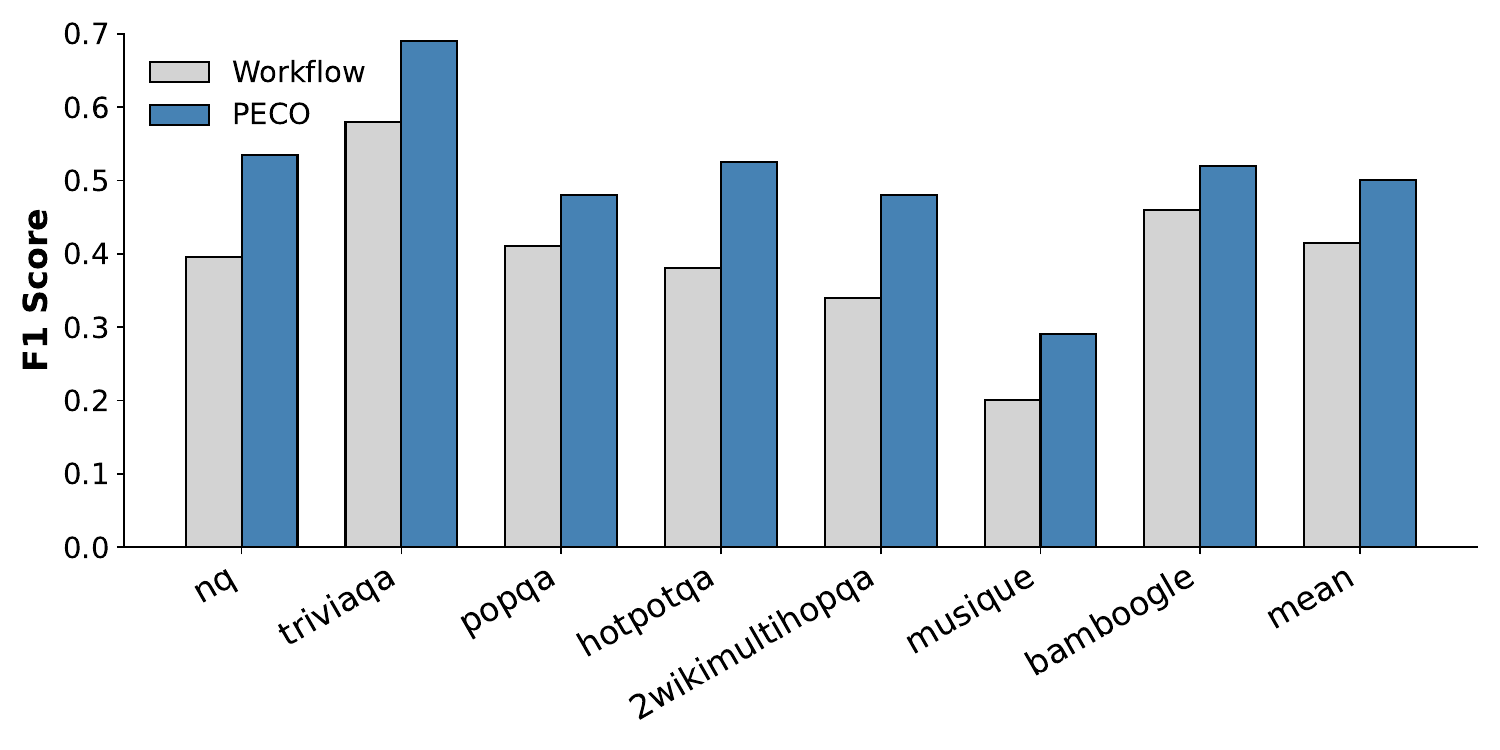}
  \caption{Ablation study comparing planner–executor co-optimization and a simple workflow.}
  \Description{}
  \label{fig:workflow_peco}
\end{figure}

As shown in Figure~\ref{fig:workflow_peco}, the results clearly demonstrate the critical role of our PECO training. On average, our full PECO-trained model achieves an F1-score of 0.50, a substantial 8 percentage point improvement over the prompt-guided baseline's score of 0.42.

This performance gap is particularly evident on complex multi-hop QA datasets. For instance, on 2WikiMultiHopQA, PECO boosts the F1-score from approximately 0.34 to 0.48. Similarly, on Musique, the score improves from 0.20 to 0.29. This indicates that while the hierarchical structure provides a solid foundation, it is the reinforcement learning process that enables the agent to master the sophisticated strategies required for multi-step reasoning and evidence refinement. The experimental result underscores that static prompting is not enough. Through outcome-supervised \textbf{RL}, the agent learns nuanced and optimized policies for both planning and execution, developing a true coordination between the two roles that goes far beyond what prompt-following can achieve.

\subsubsection{Reward Function Ablation} 
\begin{figure}[t]
  \centering
  \includegraphics[width=0.8\columnwidth]{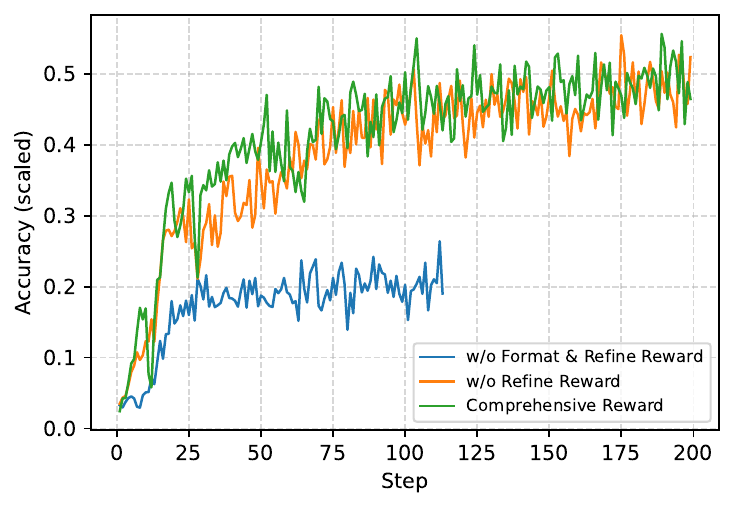}
  \caption{Performance comparison over training steps.}
  \Description{}
  \label{fig:reward_ablation_training_curve}
\end{figure}

To dissect the contribution of each component in our composite reward function (Section 3.3.4), we conducted a detailed ablation study, shown in Figure~\ref{fig:reward_ablation_training_curve} and Table~\ref{tab:exp_ablation_reward}.

Training the agent with only the final answer reward resulted in severe reward hacking; the agent would often answer directly bypassing the search tool and leaving the executor idle, leading to low accuracy and performance stagnation. Introducing a format compliance reward yielded a significant improvement. This component proved crucial for enforcing the structured plan-then-execute protocol. Furthermore, inspired by AutoRefine~\cite{Autorefine}, we added a refinement reward to incentivize effective information distillation. This final model achieved the highest performance, highlighting the importance of rewarding both correct procedures and information processing quality. 

\begin{table}[t]
    \centering
    \caption{
    Ablation study over different reward contributions in CoDA. (F1)}
    \resizebox{\linewidth}{!}{
    \begin{tabular}{lcccccccc}
        \toprule
        & \multicolumn{3}{c}{Single-Hop QA} & \multicolumn{4}{c}{Multi-Hop QA}       & \multicolumn{1}{l}{} \\
        \cmidrule(lr){2-4} \cmidrule(lr){5-8}
        Model Variants            & NQ      & TriviaQA   & PopQA   & HotpotQA & 2wiki & Musique & Bamboogle & Avg.                 \\
        \midrule
        \rowcolor[gray]{0.9}
        CoDA-Base & \textbf{0.535} & \textbf{0.688} & \textbf{0.496} & \textbf{0.525} & \textbf{0.480} & \textbf{0.289} & \textbf{0.518} & \textbf{0.500} \\
        \quad{w/o} Refine Reward & 0.505 & 0.662 & 0.501 & 0.481 & 0.373 & 0.215 & 0.427 & 0.452 \\
        \quad{w/o} Format Reward \& Refine & 0.187 & 0.331 & 0.147 & 0.188 & 0.187 & 0.085 & 0.200 & 0.190 \\
        \bottomrule
        \end{tabular}
    }
    \label{tab:exp_ablation_reward}
\end{table}

\begin{table}[t]
    \centering
    \small
    \caption{
    Performance comparison between AutoRefine and CoDA on more than 3 hops questions (Top-3 retrieval)
    }
    \resizebox{0.8\linewidth}{!}{
    \begin{tabular}{lccccc}
        \toprule
        & & \multicolumn{2}{c}{3, 4 hops} \\
        \cmidrule(lr){3-4} 
        Model & Metric & 2wiki & Musique & Avg. & \textbf{Improve \%} \\
        \midrule
        \rowcolor[gray]{0.9}
        \multicolumn{6}{l}{\textbf{Qwen2.5-3B-Base}} \\
        \multirow{3}{*}{AutoRefine} 
            & EM & 0.452 & 0.060 & 0.256 & --\\
            & F1 & 0.464 & 0.138 & 0.301 & --\\
            & CEM & 0.452 & 0.078 & 0.265 & --\\
        \midrule
        \multirow{3}{*}{CoDA} 
            & EM & \textbf{0.516} & \textbf{0.106} & \textbf{0.311} & \textbf{+21.48\%}\\
            & F1 & \textbf{0.523} & \textbf{0.197} & \textbf{0.360}  & \textbf{+19.60\%}\\
            & CEM & \textbf{0.516} & \textbf{0.137} & \textbf{0.326} &  \textbf{+23.02\%} \\
        \bottomrule
    \end{tabular}
    }
    \label{tab:34hop}
\end{table}

\begin{table*}[ht!]
    \caption{Case Study. The predictions are colored \colorIncorrect{Red} if the model answers incorrectly and \colorCorrect{Green} if it answers correctly. The core evidences are colored \colorEvidence{Blue}.}
    \vspace{-.5em}
    \label{tab:case_study}
    \small
    \resizebox{\linewidth}{!}{
    \begin{tabular}{p{0.98\linewidth}}
    \toprule\hline
    \textbf{Question: } Where do greyhound buses leave from the in the city where the headquarters of the production company which produced A Cosmic Christmas is located?  \\
    \textbf{Ground Truth:} \colorCorrect{Toronto Coach Terminal}
    \\ 
    \textbf{AutoRefine:}
    \\
    \token{think} To answer the question "Where do greyhound buses leave from the city where the headquarters of the production company which produced A Cosmic Christmas is located?", I'll issue a search request.\token{/think} 
    \\
    \token{search} Where do greyhound buses leave from the city where the headquarters of the production company which produced A Cosmic Christmas is located \token{/search} \token{documents}[Doc 1: Williams Street]...[Doc 2: Transportation Manufacturing Corporation]...[Doc 3: Liverpool Santa Dash] ... [Doc 19: The Big Bus] [Doc 20: Atlantic Greyhound Lines]...\token{/documents}
    \\
    \token{refine} From the search results, I know that the production company which produced A Cosmic Christmas is Sony Pictures Animation. To find out the city where the headquarters of Sony Pictures is located, I'll issue another search request.\token{/refine} c
    \\
    \token{search} Where is the headquarters of Sony Pictures located  \token{/search} \token{documents}[Doc 1: Sony Pictures] Sony Pictures Sony Pictures Entertainment, Inc. ... Based in \colorIncorrect{Culver City, California}, it encompasses Sony's motion picture, television production and distribution. [Doc 2: Sony Pictures Studios]...[Doc 3: History of Sony] ... [Doc 19: Sony Pictures Studios] [Doc 20: Sony]...\token{/documents}
    \\
    \token{refine} From the search results, I know that the headquarters of Sony Pictures is located in Culver City, California. To find out where greyhound buses leave from the city where the headquarters of Sony Pictures is located, I'll issue another search request.\token{/refine}
    \\
    \token{search}  Where do greyhound buses leave from the city Culver City, California\token{/search}
    \token{documents}
    [Doc 1: Culver CityBus]...
    [Doc 10: Greyhound Lines](San Jose and Oakland). A stop in the city of San Francisco was added in December 2013 along with a new route between Los Angeles and Las Vegas.
    [Doc 20: Transportation in California]...
    \token{/documents}
    \\
    \token{refine} From the search results, I know that greyhound buses leave from Culver City, California.\token{/refine}
    \token{answer} \colorIncorrect{Culver City} \token{/answer}.
    \\

    \textbf{CoDA:}
    \\
    (\textbf{Planner Trajectory})
    \\
    \token{think} I need to find out where Greyhound buses leave from the city where the headquarters of the production company that produced A Cosmic Christmas is located. \token{/think}
    \\
    \token{task} What is the production company that produced A Cosmic Christmas? \token{/task}
    \\
    \token{result} \colorEvidence{Nelvana} \token{/result}
    \\
    \token{task} What is the city where the headquarters of Nelvana is located?  \token{/task}
    \\
    \token{result}\colorEvidence{Toronto, Ontario} \token{/result}
    \\
    \token{task} Where do Greyhound buses leave from Toronto, Ontario? \token{/task}
    \\
    \token{result}\colorEvidence{Toronto Coach Terminal}\token{/result}
    \\
    \token{answer} \colorCorrect{Toronto Coach Terminal} \token{/answer}
    \\ 
    (\textbf{Executor Trajectory}) 
    \\
    \textbf{Executor 1}
    \token{task} What is the production company that produced A Cosmic Christmas? \token{/task}
    \\
    \token{think} I need to find out the production company that produced A Cosmic Christmas. \token{/think}
    \\
    \token{search} production company that produced A Cosmic Christmas \token{/search}
    \token{documents}[Doc 1: A Cosmic Christmas] A Cosmic Christmas A Cosmic Christmas is one of the first productions made by \colorEvidence{Nelvana}. It premiered on December 4, 1977, in Canada on CBC Television. ... [Doc 2: A Cosmic Christmas]...[Doc 3: Christmas on Mars] ... [Doc 19: A Cosmic Christmas] [Doc 20: Cosmos: A Spacetime Odyssey]...\token{/documents}
    \\
    \token{refine} Based on the documents, the production company that produced A Cosmic Christmas is Nelvana. \token{/refine}
    \\
    \token{result} \colorEvidence{Nelvana} \token{/result}
    \\
    ...
    \\

    \textbf{Executor 3}
    \token{task} Where do Greyhound buses leave from Toronto, Ontario? \token{/task}
    \\
    \token{think} I need to find out where Greyhound buses leave from Toronto, Ontario. \token{/think}
    \\
    \token{search} Greyhound buses leaving from Toronto, Ontario \token{/search}
    \token{documents}
    [Doc 1: A Greyhound Canada] 
    [Doc 2: Greyhound Canada]...  A list of cities served by this service: NeOn, ...
 is a service operated by Greyhound Canada in cooperation with Trailways of New York and Greyhound Lines between the New Yorker Hotel in Manhattan and \colorEvidence{the Toronto Coach Terminal}....
    [Doc 3: Bus companies in Ontario]
    ...
    [Doc 19: Bus companies in Ontario] 
    [Doc 20: Greyhound Canada]
    \token{/documents}
    \\
    \token{refine} Based on the documents, Greyhound buses leave from the Toronto Coach Terminal in Toronto, Ontario.\token{/refine}
    \\
    \token{result} \colorEvidence{Toronto Coach Terminal } \token{/result}
    \\
    
    \hline\bottomrule
    \end{tabular}
    }
\end{table*}

\subsection{Effect Analysis}
\subsubsection{Long-Context Robustness Test (RQ3)}
\label{exp:long_context}
To evaluate our framework's resilience against "context explosion", we conducted a robustness test simulating information overload. We systematically increased the volume of potentially relevant information by varying the search tool's top-k parameter (3 to 30), returning 200-token document chunks (up to 4K–6K tokens per call).

\begin{figure}[t]
  \centering
  \includegraphics[width=\columnwidth]{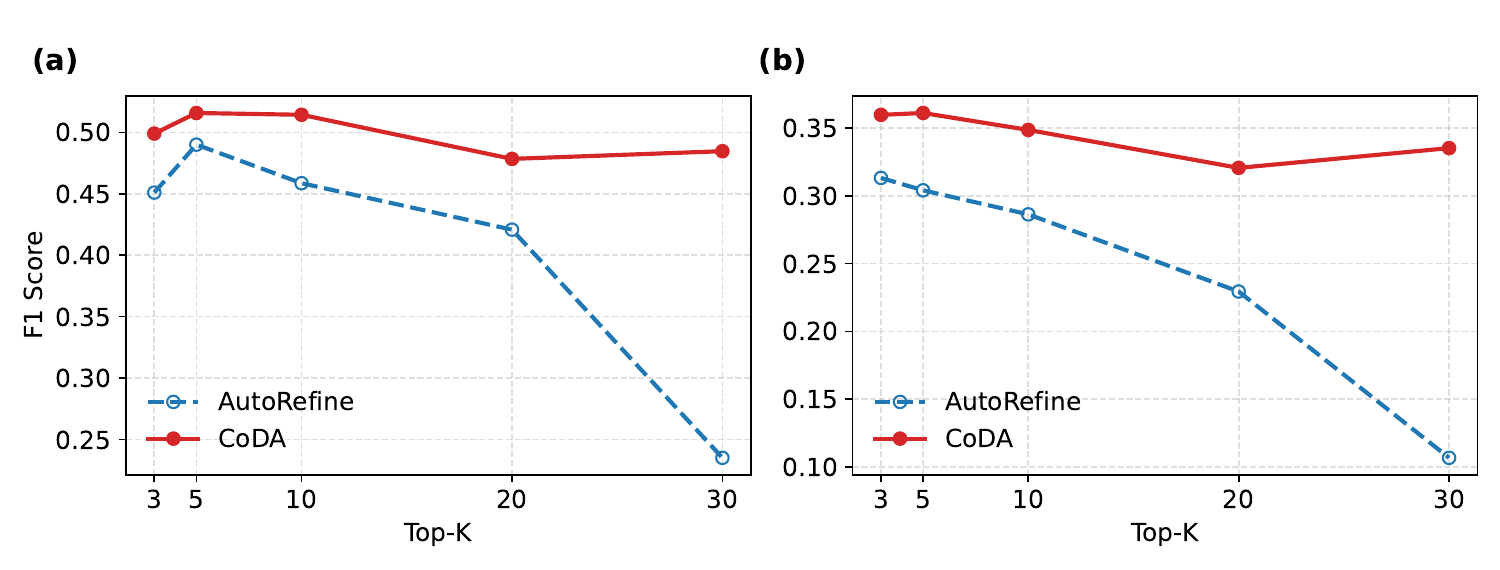}
  \caption{Performance robustness (a) across all datasets (b) multi-hop($\geq3$) questions as context length (Top-K) increases.}
  \label{fig:topk}
\end{figure}

Figure~\ref{fig:topk}a demonstrates CoDA's superior robustness. The performance of the AutoRefine baseline shows a significant and steady decline as the number of retrieved documents increases. Its F1-score falls from 0.49 to 0.24, a 52\% relative decrease, highlighting its vulnerability to cognitive overload.

In striking contrast, CoDA's performance remains remarkably stable across all $k$ values. This resilience is a direct consequence of our context-decoupled design. Because its workspace is ephemeral and its output is a condensed summary, the information overload is effectively contained and resolved at a low level, preventing pollution of the critical high-level planning process. It also aligns with our complexity analysis in Section \ref{sec:appendix_complexity}: the peak complexity of a monolithic agent scales as $\mathcal{O}((H \cdot L_{\text{doc}})^2)$, while CoDA scales as $\mathcal{O}(\max((H \cdot L_{\text{res}})^2, (h \cdot L_{\text{doc}})^2)$, avoiding quadratic growth in the number of hops $H$ when $L_{\text{res}} \ll L_{\text{doc}}$ and $h$ is small.

\subsubsection{Multi-hop Long-Context Settings}
The challenge of context overload becomes particularly severe in multi-hop QA, where information from sequential retrieval steps accumulates and amplifies noise. We evaluate both CoDA and monolithic baselines on a set of 3,916 artificially selected questions requiring more than three reasoning hops, drawn from 2WikiMultiHopQA and Musique. As shown in Table~\ref{tab:34hop}, our method performs significantly better than AutoRefine, achieving average relative improvements of +21.48\% in EM, +19.60\% in F1, and +23.02\% in CEM across the two multi-hop datasets (3--4 hops), thereby demonstrating the effectiveness of the hierarchical structure in tackling complex multi-hop questions. 

Figure~\ref{fig:topk}b shows that the performance of monolithic agents like AutoRefine~\cite{Autorefine} exhibits a dramatic performance drop, as their context is increasingly polluted by raw documents retrieved at each hop. In contrast, CoDA demonstrates remarkable resilience. Its modular architecture effectively quarantines the noisy, long-context documents within the Executor's ephemeral workspace. The \emph{Executor} processes raw documents in an isolated workspace and outputs only concise summaries, while the \emph{Planner} maintains a clean, high-level strategic context that is never directly exposed to these unprocessed inputs. This decoupling prevents information dilution and preserves reasoning quality. As a result, CoDA maintains stable and robust performance even when faced with extensive context on top of multi-hop reasoning challenges, demonstrating clear architectural advantages in complex, long-context, multi-hop scenarios where conventional approaches fail.

\subsection{Case Study}
As shown in Table~\ref{tab:case_study}, we present a case study on a complex multi-hop question from Musique, which requires: (1) identifying a film’s production company, (2) determining the city of its headquarters, and (3) finding the Greyhound bus station in that city. AutoRefine issues a single broad query covering the entire task, returning noisy and ambiguous results. This leads to an early hallucination—incorrectly identifying \emph{Sony Pictures Animation} as the production company—which sends the agent down a wrong reasoning path and yields an incorrect final answer.

In contrast, CoDA’s hierarchical structure demonstrates superior robustness. The Planner methodically decomposes the problem into three distinct, manageable sub-tasks: (1) identify the production company, (2) find its headquarters' city, and (3) find the bus station in that city. Each sub-task is delegated to an Executor sequentially, which operates in a clean, isolated context. As shown in the Executor’s trace, this isolation allows it to formulate a simple, precise search query ("production company that produced A Cosmic Christmas"), which yields the correct entity, "Nelvana". It clearly shows that the accumulation of unfiltered, verbose information in a single context window pollutes the reasoning process, making the monolithic agent prone to early-stage errors, thus leading to final failure. However, CoDA’s design principle of context decoupling directly leads to reliable task-decomposition and step-by-step reasoning.

\section{Conclusion}

In this paper, we presented CoDA, a backbone-shared Planner–Executor framework that mitigates context explosion by decoupling high-level planning from low-level tool use. Trained end-to-end using our proposed Planner-Executor Co-Optimization (PECO) scheme, CoDA achieves significant performance improvements over state-of-the-art baselines across seven QA benchmarks.

Despite these promising results, we acknowledge certain limitations in our current evaluation. While Qwen2.5-3B validates our core hypothesis regarding the efficacy of context decoupling, scaling this framework to larger foundation models (e.g., 70B+) remains to be verified to ensure consistent gains. Furthermore, our experiments focused primarily on open-domain QA. Extending CoDA to broader agentic tasks is a critical direction for future work. Specifically, we aim to validate CoDA's robustness in complex decision-making environments such as \textbf{GAIA}, \textbf{SWE-bench}, and \textbf{WebArena}. Finally, we plan to investigate nested CoDA architectures to address reasoning chains exceeding five hops, balancing accuracy with system complexity.

\newpage

\bibliographystyle{ACM-Reference-Format}

\bibliography{references}

\end{document}